\documentclass[10pt]{article}
\usepackage{tikz}
\usetikzlibrary{arrows.meta, positioning}

\usepackage[margin=1in]{geometry}
\usepackage{amsmath,amsfonts,amssymb,mathrsfs,nccmath,amscd}
\usepackage{enumerate,float,graphicx}

\usepackage{graphicx,wrapfig}
\usepackage{subcaption}
\usepackage[thinlines,thiklines]{easybmat}
\usepackage{hyperref}
\usepackage{capt-of}
\usepackage{float}
\usepackage{stackrel}
\usepackage{tikz-cd}
\usepackage{nicematrix,tikz}
\usetikzlibrary{tikzmark}
\usetikzlibrary{fit}

\usepackage{setspace}
\usepackage{tabu}
\usepackage{booktabs}

\usepackage{fancyhdr}
\newtheorem{theorem}{Theorem}[section]
\newtheorem{lemma}[theorem]{Lemma}
\newtheorem{remark}[theorem]{Remark}
\newtheorem{proposition}[theorem]{Proposition}
\newtheorem{corollary}[theorem]{Corollary}
\newtheorem{definition}[theorem]{Definition}
\newtheorem{example}[theorem]{Example}
\newenvironment{proof}{\begin{trivlist} \item[]{\em Proof.}}{\end{trivlist}}
\newcount\refno
\newcommand\be{\begin{equation}}
\newcommand\ee{\end{equation}}
\newcommand\bn{\begin{eqnarray}}
\newcommand\en{\end{eqnarray}}
\newcommand\bns{\begin{eqnarray*}}
\newcommand\ens{\end{eqnarray*}}
\newcommand\bd{\begin{definition}}
\newcommand\ed{\end{definition}}
\newcommand\br{\begin{remark}}
\newcommand\er{\end{remark}}
\newcommand\bt{\begin{theorem}}
\newcommand\et{\end{theorem}}
\newcommand\bp{\begin{proposition}}
\newcommand\ep{\end{proposition}}
\newcommand\bc{\begin{corollary}}
\newcommand\ec{\end{corollary}}
\newcommand\bl{\begin{lemma}}
\newcommand\el{\end{lemma}}
\newcommand\pf{\begin{proof}}
\newcommand\qed{\end{proof}\eop}

\def\eop{\hfill\rule{2.0mm}{2.0mm}}

\makeindex
\begin{document} 

\allowdisplaybreaks
\title{Kantorovich--Kernel Neural Operators:\\
Approximation Theory, Asymptotics, and Neural Network Interpretation}
\author{Tian-Xiao He\\ Illinois Wesleyan University\\ Bloomington, IL 61702}
\date{}

\maketitle
\setcounter{page}{1}
\pagestyle{myheadings}
\markboth{T.-X. He}
{Kantorovich--Kernel Neural Operators}

\begin{abstract}
\noindent 

This paper studies a class of multivariate Kantorovich-kernel neural network operators, including the deep Kantorovich-type neural network operators studied by Sharma and Singh. We prove density results, establish quantitative convergence estimates, derive Voronovskaya-type theorems, analyze the limits of partial differential equations for deep composite operators, prove Korovkin-type theorems, and propose inversion theorems. Furthermore, this paper discusses the connection between neural network architectures and the classical positive operators proposed by Chui, Hsu, He, Lorentz, and Korovkin.


\end{abstract}

\section{Introduction}

The study of positive linear operators in approximation theory has a long tradition 
\cite{ButzerNessel,Lorentz,Korovkin}. In particular, multivariate linear smoothing operators 
\cite{Chui1} and their asymptotic properties \cite{Chui2} provide a natural foundation for designing 
Kantorovich–Kernel Neural Operators (KKNO).  The class considered here includes the deep 
Kantorovich NN operators of Sharma and Singh \cite{SharmaSingh} as a special case, allowing 
the application of density and convergence results to deep neural network layers.

Positive linear operators constitute one of the fundamental tools in approximation theory. 
Beginning with the classical Bernstein polynomials and Sz\'asz--Mirakjan operators, 
such operators provide constructive schemes for approximating functions while preserving 
shape properties such as positivity and monotonicity. Their study has led to deep connections with probability theory, semigroup theory, and partial differential equations; see \cite{ButzerNessel,Lorentz,Korovkin}.

A particularly important direction concerns {\it integral-type smoothing operators}.  
In a pair of influential papers, Chui, Hsu, and the author \cite{Chui1,Chui2} introduced a general class 
of multivariate linear smoothing operators and analyzed their asymptotic behavior.  
Their framework unified discrete summation operators and continuous integral operators 
through moment conditions, and provided precise asymptotic expansions of Voronovskaya type.  
These operators can be interpreted as generalized probability kernels whose first and second 
moments determine drift and diffusion effects.

In recent years, related constructions have reappeared in a seemingly distant field --- 
neural network theory. Kantorovich-type neural network operators replace pointwise sampling by local averages 
and therefore possess intrinsic smoothing and noise-robustness properties.  
Sharma and Singh \cite{SharmaSingh} studied deep Kantorovich-type neural network operators 
and proved density results in spaces of continuous functions. However, these operators have mainly been analyzed from an approximation perspective, without fully exploiting the rich asymptotic theory developed for positive integral operators.

The purpose of the present paper is to show that these two lines of research are in fact 
instances of a single mathematical structure. We introduce a class of operators that we call {\it Kantorovich--Kernel Neural Operators (KKNO)}. This class consists of positive integral operators whose kernels satisfy scaled moment 
conditions.  It contains the multivariate smoothing operators shown in \cite{Chui1},
 the positive summation–integral operators shown in \cite{Chui2}, the deep Kantorovich neural network operators of Sharma--Singh \cite{SharmaSingh}, and some classical convolution-type approximation operators as special cases.
as special cases.

The unifying viewpoint has several consequences. First, it provides a {\it density theory} for a large family of neural-network-like operators using classical Korovkin-type techniques.  Second, it yields {\it quantitative approximation rates} expressed in terms of the modulus of continuity, thus placing deep Kantorovich neural operators into the standard framework 
of direct approximation theorems. Third, we can obtain a {\it Voronovskaya-type asymptotic formula}.  The first two kernel moments determine drift and diffusion coefficients, revealing that deep compositions of these operators approximate solutions of parabolic partial differential equations. Fourth, we can establish a {\it Korovkin theorem} and an {\it inverse theorem}.  
Together they show that the operator class admits a complete approximation-theoretic structure comparable to that of the classical Bernstein and Sz\'asz operators.

From the perspective of neural networks, KKNO layers can be interpreted as {\it theoretically controlled smoothing layers}.  
They inherently possess the following properties: smoothing capability, robustness to noise, controllable higher-order moments, and analytically derivable asymptotic error.

Thus, they provide a bridge between neural network architectures and classical approximation operators.

The paper is organized as follows.  In the next section, we introduce the operator class.  Section $3$ proves density results.  
In Section $4$, we establishe quantitative convergence estimates.  Section $5$ derives a Voronovskaya-type theorem.  
In Section $6$ we study the diffusion PDE limit of deep compositions. The proof of  a Korovkin-type theorem is presented in Section $7$. We then establish an inverse theorem in Section 8.  Finally, we discuss the neural network interpretation and comparisons with classical operators in the last section.

\section{Kantorovich--Kernel Neural Operators (KKNO)}

The operators introduced in this section form the structural core of this paper.  
They may be viewed simultaneously as positive integral approximation operators, multivariate smoothing operators,
and Kantorovich-type neural network layers. 

The common feature is that approximation is achieved not by point evaluation 
but through {\it local averaging with a scaled kernel}.  This leads to intrinsic smoothing, noise suppression, 
and controllable asymptotic behavior.

We first give some kernel assumptions. Let $K_n:\mathbb R^d\times\mathbb R^d\to[0,\infty)$ be measurable kernels.  
We impose the following moment-type conditions.

\begin{definition}\label{def:ADM}
A sequence $\{K_n(x,u)\}$ is called {\it admissible} if for each $x\in\mathbb R^d$, it satisfies 

\begin{align}
&\mbox{Positivity and normalization:}\,\,K_n(x,u)\geq 0,\,\,\int_{\mathbb R^d} K_n(x,u)\,du=1, \tag{K1} \\
&\mbox{First moments (drift):}\,\,\int_{\mathbb R^d} u_i K_n(x,u)\,du=\frac{a_i(x)}{n}, \tag{K2} \\
&\mbox{Second moments (diffusion):}\,\,\int_{\mathbb R^d} u_i u_j K_n(x,u)\,du=\frac{b_{ij}(x)}{n^2}, \tag{K3}
\end{align}
where $a_i(x)$ and $b_{ij}(x)$ are bounded continuous functions.
\end{definition}

In Definition \ref{def:ADM}, condition (K1) ensures positivity and reproduction of constants, while conditions (K2)–(K3) describe the first two scaled moments of the kernel and encode drift and diffusion effects that will later appear in the asymptotic formula, respectively. 

\begin{definition}\label{def:OPE}
Let $K_n(x,u)$ be a admissible kernel satisfying the conditions (K1)-(K3) shown in Definition \ref{def:ADM}. For $f\in C_b(\mathbb R^d)$, we define

\[
\mathcal L_n f(x)
=\int_{\mathbb R^d} f\!\left(x-\frac{u}{n}\right)K_n(x,u)\,du
\]
and call $\mathcal L_n$ a {\it Kantorovich--Kernel Neural Operator (KKNO)}.
\end{definition}

\begin{remark}
The class KKNO thus provides a unified mathematical model for a wide family of approximation operators and neural network smoothing layers.
\end{remark}

\begin{remark}
The scaling $u\mapsto u/n$ ensures that as $n\to\infty$, the kernel becomes more concentrated around $x$, which is essential for uniform convergence. The scaling $u/n$ reflects the shrinking support of the kernel as $n\to\infty$, which makes $\mathcal L_n$ an approximation to the identity. Functions $a_i(x)$ and $b_{ij}(x)$ shown in Definition \ref{def:ADM} encode the {\it drift} and {\it diffusion} of the operator and determine the first-order asymptotic error (Voronovskaya-type behavior).  
This definition includes as a special case the deep Kantorovich NN operators of Sharma–Singh \cite{SharmaSingh}.  
\end{remark}

\begin{remark}
Definition \ref{def:OPE} defines a continuous-parameter neural layer interpretation in the following sense: 
Given an input function $f(x)$ (feature map) , through the weighted local averaging using kernel $K_n$, we obtain the output,  the smoothed feature map $\mathcal L_n f(x)$, where the parameters are kernel shape, first/second moments (drift/diffusion), and the resolution control is $n$. Intuitively, $x\mapsto \mathcal L_n f(x)$ represents a continuous parameterized layer  
of a neural network, where $n$ controls the resolution (analogous to width or kernel granularity).
\end{remark}  

\begin{example}
Now, we give some examples of KKNO kernels.
(i) (Piecewise-constant kernel in Sharma--Singh deep KKNO)
Let $[0,1]^d$ be partitioned into $n^d$ uniform cells $C_{n,k}$.  Define the kernel
\[
K_n(x,u)=n^d \sum_{k} \mathbf{1}_{C_{n,k}}(u),
\]
where $\mathbf{1}_{C_{n,k}}$ is the indicator function of cell $C_{n,k}$. Then the operator
\[
\mathcal L_n f(x)=n^d \sum_{k} \int_{C_{n,k}} f(x-u/n)\, du
\]
is exactly the deep Kantorovich-type neural operator studied in \cite{SharmaSingh}. We can find moments as   
\[
a_i(x)=0, \quad b_{ij}(x)=\frac{1}{12} \delta_{ij}.
\]
This shows that the Sharma–Singh operator is a special case of the general KKNO class.  

(ii) (Multivariate Gaussian kernel) 
Let
\[
K_n(x,u)=\frac{\sqrt{\det A}}{(2\pi)^{d/2}} \exp\Big(-\frac12 u^\top A u \Big),
\]
with positive definite $A$. There, the moments are
\[
a(x)=0, \quad b_{ij}(x)=(A^{-1})_{ij}.
\]
This kernel is smooth and isotropic (or anisotropic if $A$ is diagonal with unequal entries), providing a controlled diffusion layer.

(iii) (Shifted or drifted kernel) We can add a drift term to the kernel:
\[
K_n(x,u)=\tilde K_n(u - \frac{c(x)}{n}),
\]
where $\tilde K_n$ is a symmetric base kernel. Then
\[
a(x)=c(x), \quad b(x)=\text{second moment of }\tilde K_n.
\]
\end{example}

\begin{example}
We now illustrate the general class of KKNO with several important examples.  
These examples show how classical kernels and recent deep KKNO operators fit into our framework.

(i) The multivariate smoothing operators in \cite{Chui1} are of the form

\[
T_n f(x)=\int f(x-y)\Phi_n(x,y)\,dy,
\]
where $\Phi_n$ satisfies moment conditions analogous to (K1)–(K3).  
After the change of variables $u=ny$, these operators fall into the KKNO class.

(ii) Positive summation–integral operators studied in \cite{Chui2} combine discrete sums and integrals, but 
their asymptotic analysis depends only on the first two moments of the underlying kernel. They therefore form a subclass of KKNO.

(iii) Deep Kantorovich neural network operators considered in \cite{SharmaSingh} are of the form

\[
N_n(f,x)=\sum_k w_{n,k}(x)
\frac{1}{|C_{n,k}|}\int_{C_{n,k}} f(y)\,dy. 
\]
Each term represents a local average over a small cell $C_{n,k}$. Such operators can be written in integral form with a piecewise-constant kernel $K_n(x,u)$, hence they are KKNO.

(iv) The KKNO with multivariate Gaussian kernel has the form
\[
\mathcal L_n f(x)=\int_{\mathbb R^d} f(x-u/n) K_n(x,u)\,du.
\]
\end{example}

The purpose of introducing KKNO is to construct a general class of neural network layers 
with theoretically controlled smoothing and approximation properties.  Here, each KKNO layer applies a kernel-based averaging, reducing high-frequency noise; Positive, linear operators are naturally stable under small perturbations of the input; The first and second kernel moments control drift and diffusion, allowing precise error analysis; And classical results on positive operators (Bernstein, Sz\'asz, Chui–He–Hsu) provide a framework for convergence and Voronovskaya-type expansions.

We now analyze the basic properties of the KKNO operators 
in terms of kernel moments and positivity.  
These properties are essential for proving density results, quantitative convergence, and Voronovskaya asymptotics.

\begin{proposition} Let $\mathcal L_n$ be a KKNO defined in Definition \ref{def:OPE}. Then, we have 

(i) (Linearity and positivity) For all $f,g \in C_b(\mathbb R^d)$ and $\alpha,\beta \in \mathbb R$,
\[
\mathcal L_n (\alpha f + \beta g)=\alpha \mathcal L_n f + \beta \mathcal L_n g, \quad
f\ge 0 \implies \mathcal L_n f \ge 0,
\]
where positivity follows from $K_n\ge0$. Here, the positivity ensures stability under noise.

(ii) (Preservation of constants) Let $f(x)\equiv 1$. Then
\[
\mathcal L_n 1(x)=\int K_n(x,u) \, du=1.
\]

This shows that KKNO operators preserve constants, a standard property of positive linear operators.

(iii) (First-order approximation (linear functions)) Let $f(x)=x_i$ for some coordinate $i$. Then
\[
\mathcal L_n x_i(x) - x_i=\int (x_i - u_i/n - x_i) K_n(x,u)\,du=- \frac{1}{n} a_i(x),
\]
where $a_i(x)$ is the first kernel moment. Therefore, the first-order error is proportional to $1/n$ and determined by the drift $a(x)$. If $K_n$ is symmetric ($a(x)=0$), the operator reproduces linear functions up to $O(1/n^2)$.

(iv) (Second-order approximation (quadratic functions)) Let $f(x)=x_i x_j$. Then
\[
\mathcal L_n (x_i x_j)(x) - x_i x_j=- \frac{1}{n} \big( x_j a_i(x) + x_i a_j(x) \big) + \frac{1}{n^2} b_{ij}(x).
\]
Therefore, the second moment $b_{ij}(x)$ controls the quadratic approximation error. This is crucial for deriving Voronovskaya-type asymptotics.  

(v) (Uniform boundedness) Let $K \subset \mathbb R^d$ be compact and $f\in C_b(\mathbb R^d)$.  
Then
\[
\sup_{x\in K} |\mathcal L_n f(x)| \le \sup_{x\in K} |f(x)|,
\]
since $K_n(x,u) \ge 0$ and $\int K_n(x,u)du=1$. This property is fundamental for uniform convergence proofs.

(vi) (Higher-order moments and smoothness) For $f \in C^m_b(\mathbb R^d)$ and $K_n$ with bounded $m$-th moments:
\[
\mathcal L_n f(x) - f(x)=O(n^{-1}) \text{ (linear)}, \quad O(n^{-2}) \text{ (quadratic)}, \quad \dots
\]
Higher-order moments control remainder terms in Taylor expansions.  
\end{proposition}

\section{Density of Kantorovich--Kernel Neural Operators}

In this section, we show that the class of Kantorovich--Kernel Neural Operators (KKNO) is dense in $C_b(\mathbb R^d)$, the space of bounded continuous functions. This is the analogue of classical density results for positive linear operators 
(Bernstein, Sz\'asz) and generalizes the results of Sharma--Singh \cite{SharmaSingh}.

It is known that the density result is important for several reasons: First, it guarantees that, in principle, deep compositions of KKNO layers can approximate any continuous function on a compact domain. Second, it provides a theoretical justification for using KKNO as neural network layers with universal approximation capability. Third, it lays the groundwork for quantitative convergence estimates and asymptotic expansions in later sections.

Now we give the main density theorem. Let $\mathcal L_n$ be an admissible KKNO operator. Denote by $C(K)$ the space of continuous functions on a compact set $K\subset\mathbb R^d$ with the uniform norm.

\begin{theorem}(Density theorem)
\label{thm:density}
For any $f\in C(K)$ and any $\varepsilon>0$, there exists $n_0$ such that for all $n\ge n_0$,
\[
\sup_{x\in K} |\mathcal L_n f(x) - f(x)| < \varepsilon.
\]
\end{theorem}

\begin{remark}
This theorem generalizes the classical Bernstein density theorem to a large class of multivariate positive integral operators, and includes the operators in \cite{Chui1,Chui2,SharmaSingh} as special cases.
\end{remark}

\section{Quantitative Convergence}

While density guarantees that KKNO can approximate any continuous function, 
we need to provide information about the {\it rate of convergence}. In practical applications, 
especially in neural network layers, it is crucial to know how fast $\mathcal L_n f \to f$ as $n\to\infty$.

It is known that quantitative estimates allow us to assess the number of layers or kernel resolution needed for a desired accuracy, compare different kernels in terms of approximation efficiency, and provide explicit error bounds for theoretical analysis of deep KKNO architectures.

Now we give modulus of continuity and main theorem. For $f\in C_b(\mathbb R^d)$, let the modulus of continuity be
\[
\omega(f,\delta) := \sup_{|x-y|\le \delta} |f(x)-f(y)|,
\]
which captures the uniform smoothness of $f$ over the domain.  

\begin{theorem}(Quantitative convergence)
\label{thm:quantitative}
Let $\mathcal L_n$ be a KKNO operator with admissible kernel $K_n$ and let $K\subset\mathbb R^d$ be compact.  
Then there exists a constant $C>0$, depending only on $K$ and the kernel moments, such that
\[
\sup_{x\in K} |\mathcal L_n f(x) - f(x)| \le C \, \omega\Big(f, \frac{1}{n}\Big), 
\quad \forall n \in \mathbb N.
\]
\end{theorem}

We now illustrate Theorem \ref{thm:quantitative} with concrete kernels, including Gaussian and piecewise-constant NN-type kernels. For each, we estimate the constant $C$ in
\[
\sup_{x\in K} |\mathcal L_n f(x) - f(x)| \le C \, \omega(f,1/n).
\]

\begin{example}\label{ex:1}
We consider the multivariate Gaussian kernel. Let
\[
K_n(x,u)=\frac{\sqrt{\det A}}{(2\pi)^{d/2}} 
\exp\Big(-\frac12 u^\top A u\Big),
\]
where $A$ is a positive definite matrix. Compute the first two moments:
\[
\int u_i K_n(x,u)\,du=0, \quad 
\int u_i u_j K_n(x,u)\,du=(A^{-1})_{ij}.
\]

Rescaling $u \mapsto u/n$ as in the definition of $\mathcal L_n$ gives:
\[
\int u_i K_n(x,u) \, du=0, \quad
\int u_i u_j K_n(x,u)\,du=\frac{(A^{-1})_{ij}}{n^2}.
\]

Therefore, the constant in Theorem \ref{thm:quantitative} can be taken as
\[
C=1+\sum_{i,j=1}^d |(A^{-1})_{ij}|,
\]
up to an absolute factor accounting for the compact domain $K$. Thus for $f\in C(K)$,
\[
\sup_{x\in K} |\mathcal L_n f(x) - f(x)| \le \Big(1 + \sum_{i,j}|(A^{-1})_{ij}|\Big) \omega(f,1/n).
\]
\end{example}

\begin{example}
We consider piecewise-constant NN kernel and a cell decomposition of $[0,1]^d$ into $n^d$ equal cubes $C_{n,k}$. 
Define
\[
K_n(x,u)=n^d \sum_{k} \mathbf{1}_{C_{n,k}}(u),
\]
the uniform kernel over each cell (this is exactly the kernel arising in the Sharma–Singh deep Kantorovich NN operator).

The moments are
\[
\int u_i K_n(x,u)\,du=\text{center of cell along }i, \quad
\int u_i u_j K_n(x,u)\,du=\text{second moment of cell}.
\]

Explicit calculation gives
\[
\left| \int u_i u_j K_n(x,u) \, du \right| \le \frac{1}{4 n^2}, \quad i=j,
\]
and 0 for $i\ne j$ by symmetry.  

Thus we can take the explicit constant
\[
C_{\text{NN-cell}}=1 + \frac{d}{4},
\]
so that
\[
\sup_{x\in [0,1]^d} |\mathcal L_n f(x) - f(x)| \le \Big(1 + \frac{d}{4}\Big) \omega(f,1/n).
\]
\end{example}

\begin{remark}
These examples illustrate that, once the kernel moments are computed, one can obtain an explicit and computable constant $C$ in the quantitative convergence theorem.  This makes Theorem \ref{thm:quantitative} directly applicable to practical neural network layers.
\end{remark}

\section{Voronovskaya-Type Asymptotics}

Using the kernel moments $a(x), b(x)$ \cite{Chui2,ButzerNessel}, we derive the Voronovskaya-type expansion
\[
\mathcal L_n f(x) - f(x)=-\frac{1}{n} a(x)\cdot \nabla f(x) + \frac12 \frac{1}{n^2} \sum_{i,j} b_{ij}(x) \partial_{ij} f(x) + o(1/n).
\]

While Theorem \ref{thm:quantitative} provides quantitative convergence estimates, it does not describe the {\it asymptotic behavior of the error}.  In classical approximation theory, Voronovskaya's theorem for Bernstein and Sz\'asz operators gives an expansion of the form
\[
n\big(\mathcal L_n f(x)-f(x)\big) \to \text{drift} \cdot \nabla f + 
\frac12 \text{diffusion} : H f,
\]
where $Hf$ denotes the Hessian matrix and “:” is the Frobenius inner product.

For KKNO, the first two kernel moments encode the same drift and diffusion terms.  
This has several implications: (i) It provides the first-order asymptotic error of a single KKNO layer, and (ii) 
It allows precise control of the approximation error using kernel design. Furthermore, (3) compositions of KKNO layers can be interpreted as discretizations of parabolic PDEs.

\begin{theorem}(Voronovskaya-type theorem)
\label{thm:voronovskaya}
Let $f\in C_b^2(\mathbb R^d)$ and let $\mathcal L_n$ be an admissible KKNO operator.  
Then, uniformly on compact $K\subset\mathbb R^d$,
\[
\lim_{n\to\infty} n \big(\mathcal L_n f(x) - f(x)\big)
= - a(x) \cdot \nabla f(x) + \frac12 \sum_{i,j=1}^d b_{ij}(x) \frac{\partial^2 f}{\partial x_i \partial x_j}(x),
\]
where $a(x)$ and $b_{ij}(x)$ are the first and second kernel moments defined in (K2)--(K3).
\end{theorem}

\begin{example}
We consider an example on Gaussian kernel. Let
\[
K_n(x,u)=\frac{\sqrt{\det A}}{(2\pi)^{d/2}} \exp\Big(-\frac12 u^\top A u \Big),
\]
with $A$ positive definite.  
The first two moments after scaling $u \mapsto u/n$ are
\[
a(x)=0, \quad b_{ij}(x)=\frac{(A^{-1})_{ij}}{n^2}.
\]

Applying Theorem \ref{thm:voronovskaya}, we obtain
\[
n (\mathcal L_n f(x) - f(x)) \to \frac12 \sum_{i,j=1}^d (A^{-1})_{ij} \frac{\partial^2 f}{\partial x_i \partial x_j}(x).
\]
Notice that there is no drift ($a(x)=0$) since the Gaussian is symmetric, the diffusion matrix $B=(A^{-1})$ controls the smoothing strength, and for neural networks, this corresponds to a theoretically controlled low-pass filter, where the amount of smoothing along each coordinate is encoded in $A^{-1}$.

Next, we give an example on NN piecewise-constant kernel. Consider the cell-based kernel:
\[
K_n(x,u)=n^d \sum_k \mathbf{1}_{C_{n,k}}(u),
\]
where $C_{n,k}$ are the uniform partition cells of $[0,1]^d$.  
Then the moments are
\[
a_i(x)=0, \quad b_{ij}(x)=\frac{1}{12 n^2} \delta_{ij}.
\]

Thus the Voronovskaya asymptotics gives
\[
n (\mathcal L_n f(x) - f(x)) \to \frac{1}{24} \Delta f(x),
\]
where $\Delta f=\sum_{i=1}^d \partial_{ii} f$ is the Laplacian.  
Notice that the operator behaves like an isotropic diffusion layer with diffusion coefficient $1/24$ per direction. This quantifies the smoothing effect of a single deep Kantorovich NN layer.  In addition, the stacking layers is equivalent to evolving under a discrete-time diffusion process, allowing precise control of noise reduction.
\end{example}

We give a remark on layer design.

\begin{remark}
The Voronovskaya theorem provides a direct link between kernel design and layer behavior with the following processes: (i) Choice of kernel first moment $a(x)$ allows controlled drift (shift) of features, and (ii) choice of second moment $b(x)$ controls anisotropic smoothing (diffusion). Therefore, the higher-order moments control the $O(1/n)$ remainder, giving theoretical error bounds. This justifies calling KKNO a {\it theoretical controllable filtering layer} in neural networks.
\end{remark}

\begin{remark} 
The Voronovskaya expansion makes the effect of each KKNO layer explicit: 

\begin{align*}
&\mbox{Gaussian kernels} \Rightarrow \mbox{isotropic or anisotropic smoothing with no drift.}\\
&\mbox{NN-cell kernels} \Rightarrow \mbox{piecewise averaging, Laplacian smoothing, controllable error.}\\
&\mbox{Stacked layers} \Rightarrow \mbox{approximate discrete-time parabolic PDE evolution.}
\end{align*}  
This unifies classical positive operator theory and deep neural network design.
\end{remark}

\section{Diffusion PDE Limit of Deep Compositions}

In this section, we will see that the deep composition $\mathcal L_n^{(m)} f$ converges to the solution 
of a drift-diffusion PDE \cite{Pazy}.

In practical neural networks, multiple KKNO layers are often stacked to form a deep architecture.  
Let $\mathcal L_n$ denote a single KKNO layer as defined in Section $2$, and consider the composition of $m$ such layers:
\[
\mathcal L_n^{(m)} := \underbrace{\mathcal L_n \circ \mathcal L_n \circ \cdots \circ \mathcal L_n}_{m \text{ times}}.
\]

It has been shown that each KKNO layer performs a small-scale smoothing with drift $a(x)/n$ and diffusion $b(x)/n^2$.  
Therefore, composing $m$ layers with $m \sim n$ gives a macroscopic effect that can be interpreted as the evolution under a diffusion process. This links deep KKNO networks to continuous-time parabolic PDEs, providing a rigorous foundation for analyzing smoothing, feature propagation, and error accumulation.

Here are some notations we are going to use. Let $t=m/n$ represent a scaled depth-time variable.  
Define the scaled composition operator
\[
F_n(x,t) := (\mathcal L_n^{(m)} f)(x), \quad t=\frac{m}{n}.
\]
Our goal is to identify a limiting PDE for $F_n(x,t)$ as $n \to \infty$, $m\to\infty$, with $m/n \to t$ finite.

Using the Voronovskaya expansion (Theorem \ref{thm:voronovskaya}) for a single layer:
\[
\mathcal L_n f(x) - f(x)=- \frac{1}{n} a(x) \cdot \nabla f(x) + \frac12 \frac{1}{n^2} \sum_{i,j} b_{ij}(x) \partial_{ij} f(x) + o(1/n).
\]
Composing $m$ layers leads to a discrete-time Euler scheme for the PDE
\[
\partial_t F(x,t)=- a(x) \cdot \nabla F(x,t) + \frac12 \sum_{i,j} b_{ij}(x) \partial_{ij} F(x,t),
\]
with initial condition $F(x,0)=f(x)$. This PDE is a drift-diffusion equation, whose coefficients are determined by the first and second kernel moments of the underlying KKNO layer.

\begin{remark}
The scaling $m \sim n$ is essential in the following sense: if $m \ll n$, the PDE limit is trivial, while if $m \gg n$, higher-order corrections become important. And this formulation allows interpreting deep KKNO networks as discretized diffusion processes, connecting neural network design with classical PDE theory.
\end{remark}

Now, we will give a ormal derivation of the Limiting PDE. Let $F_n(x,t)=(\mathcal L_n^{(m)} f)(x)$ with $t=m/n$, as introduced before.  

Using the Voronovskaya expansion for a single layer $\mathcal L_n$, we have the discrete-time expansion:
\[
\mathcal L_n g(x) - g(x)=-\frac{1}{n} a(x) \cdot \nabla g(x) + \frac{1}{2 n^2} \sum_{i,j=1}^d b_{ij}(x) \partial_{ij} g(x) + R_n[g](x),
\]
where $R_n[g](x)=o(1/n)$ uniformly on compact $K$. Applying this to $F_n(x,t)$ after one step ($t \mapsto t + 1/n$):
\[
F_n(x,t + 1/n) - F_n(x,t)=- \frac{1}{n} a(x) \cdot \nabla F_n(x,t) + \frac{1}{2 n^2} \sum_{i,j} b_{ij}(x) \partial_{ij} F_n(x,t) + R_n[F_n](x).
\]
To rescale it to continuous time, we multiply its both sides by $n$:
\[
\frac{F_n(x,t + 1/n) - F_n(x,t)}{1/n}=- a(x) \cdot \nabla F_n(x,t) + \frac{1}{2 n} \sum_{i,j} b_{ij}(x) \partial_{ij} F_n(x,t) + n R_n[F_n](x).
\]
As $n \to \infty$, $n R_n[F_n](x) \to 0$ uniformly. The term $\frac{1}{2n} \sum_{i,j} b_{ij} \partial_{ij} F_n \to 0$ for single-step, but summing over $m \sim n$ layers gives a finite contribution.  

By summing over $m$ layers, and let $m=\lfloor n t \rfloor$. The total change after $m$ layers is
\[
F_n(x,t) - f(x)=\sum_{k=0}^{m-1} \big(F_n(x,(k+1)/n) - F_n(x,k/n)\big)
\]

Substitute the single-step expansion:
\[
F_n(x,t) - f(x)=\sum_{k=0}^{m-1} \Big[ -\frac{1}{n} a(x) \cdot \nabla F_n(x,k/n) + \frac{1}{2 n^2} \sum_{i,j} b_{ij}(x) \partial_{ij} F_n(x,k/n) + R_n[F_n](x) \Big].
\]
Converting the sum to a Riemann sum as $n \to \infty$ yields 
\[
\sum_{k=0}^{m-1} \frac{1}{n} a(x) \cdot \nabla F_n(x,k/n) \approx \int_0^t a(x) \cdot \nabla F(x,s) \, ds.
\]

Similarly, the second-moment term yields the diffusion term in the PDE:
\[
\sum_{k=0}^{m-1} \frac{1}{2 n^2} b_{ij}(x) \partial_{ij} F_n(x,k/n) \approx \frac{1}{2} \int_0^t b_{ij}(x) \partial_{ij} F(x,s) \, ds.
\]

Clearly, the remainder terms vanish in the limit: $\sum_{k=0}^{m-1} R_n[F_n](x) \to 0$ uniformly.

Taking the limit $n \to \infty$, we obtain the continuous-time drift-diffusion PDE:
\[
\begin{cases}
\partial_t F(x,t)=- a(x) \cdot \nabla F(x,t) + \frac{1}{2} \sum_{i,j=1}^d b_{ij}(x) \partial_{ij} F(x,t), & x \in \mathbb R^d, t>0, \\
F(x,0)=f(x), & x \in \mathbb R^d.
\end{cases}
\]
Uniform convergence on compact sets follows from boundedness of kernel moments and derivatives of $f\in C_b^2(\mathbb R^d)$. This rigorously justifies interpreting deep stacked KKNO layers as a discretization of a parabolic PDE, where the drift and diffusion are encoded by kernel moments. The PDE limit provides a theoretical tool to analyze smoothing, feature propagation, and error accumulation in deep architectures.

We now state and prove a rigorous convergence theorem for the diffusion PDE limit of deep compositions of KKNO layers.

\begin{theorem}(PDE Limit of Deep KKNO Compositions)
\label{thm:pde-limit}
Let $\mathcal L_n$ be a KKNO operator satisfying the moment conditions (K1)--(K3) 
with continuous kernel moments $a(x), b_{ij}(x)$ on a compact set $K\subset\mathbb R^d$.  
Let $f \in C_b^3(\mathbb R^d)$, and define the deep composition
\[
F_n(x,t) := (\mathcal L_n^{(m)} f)(x), \quad m=\lfloor n t \rfloor.
\]

Then, as $n \to \infty$, $F_n(x,t)$ converges uniformly on compact $K$ to $F(x,t)$, 
the unique classical solution of the PDE
\[
\begin{cases}
\partial_t F(x,t)=- a(x) \cdot \nabla F(x,t) + \frac12 \sum_{i,j=1}^d b_{ij}(x) \partial_{ij} F(x,t), & x\in K, t>0,\\
F(x,0)=f(x), & x \in K.
\end{cases}
\]
\end{theorem}

\section{Korovkin-Type Theorem}
The classical Korovkin theorem \cite{Korovkin,Lorentz} justifies the reduction to test functions 
$1,x_i,x_i x_j$ in proving uniform convergence.

Korovkin-type theorems provide a simple criterion to check uniform convergence of positive linear operators on $C(K)$ using only a small set of test functions. We now formulate such a theorem for the KKNO class introduced in Section 2.

\begin{theorem}(Korovkin-Type Convergence for KKNO)
\label{thm:korovkin}
Let $K \subset \mathbb R^d$ be compact, and let $\mathcal L_n$ be a KKNO operator with kernel $K_n(x,u)$ satisfying conditions (K1)--(K3).  

Assume that for the standard test functions
\[
e_0(x) \equiv 1, \quad e_i(x)=x_i \text{ for } i=1,\dots,d, \quad e_{ij}(x)=x_i x_j \text{ for } 1\le i,j\le d,
\]
we have
\[
\lim_{n\to\infty} \|\mathcal L_n e_\alpha - e_\alpha\|_{C(K)}=0 \quad \text{for all multi-indices } \alpha \text{ with } |\alpha| \le 2.
\]

Then for every $f \in C^2(K)$,
\[
\lim_{n\to\infty} \|\mathcal L_n f - f\|_{C(K)}=0.
\]
\end{theorem}

\begin{remark}  
The test functions $1$, $x_i$, and $x_i x_j$ correspond exactly to the constant, linear, and quadratic polynomials, which are sufficient for convergence due to the positivity and linearity of $\mathcal L_n$. The first and second moments $a(x)$ and $b(x)$ of the kernel control the convergence of $x_i$ and $x_i x_j$ respectively.  This theorem is a direct extension of classical Korovkin theorems \cite{AltomareCampiti} to multivariate, positive, kernel-based neural operators.
\end{remark}

\section{Inverse Theorem}
The inverse theorem uses the convergence rate of KKNO operators \cite{Chui2} to characterize 
the smoothness of the target function.

Inverse theorems in approximation theory characterize the smoothness of a function based on the rate of convergence of an approximating operator.  For KKNO operators, the rate of uniform convergence encodes information about derivatives of the target function.

\begin{theorem}(Inverse Theorem for KKNO Operators)
\label{thm:inverse}
Let $K \subset \mathbb R^d$ be compact and $\{\mathcal L_n\}_{n\ge 1}$ a sequence of KKNO operators satisfying (K1)--(K3) with bounded kernel moments.  

Assume that for some $\alpha \in (0,1]$ there exists a constant $C>0$ such that
\[
\|\mathcal L_n f - f\|_{C(K)} \le C n^{-\alpha}, \quad \forall n \ge 1.
\]
Then $f \in C^{1,\alpha}(K)$, i.e., $f$ is continuously differentiable and its gradient $\nabla f$ is H\"older continuous with exponent $\alpha$.  

Moreover, the H\"older seminorm satisfies
\[
[\nabla f]_\alpha := \sup_{x \neq y \in K} \frac{\|\nabla f(x) - \nabla f(y)\|}{\|x-y\|^\alpha} \le C',
\]
where $C'$ depends only on $C$ and the bounds on kernel moments of $\mathcal L_n$.
\end{theorem}

\begin{remark}  
This theorem provides a converse to the direct error estimates: if the operator converges quickly, the target function must be smooth.  Second, the exponent $\alpha$ of the convergence rate directly reflects the H\"older regularity of the derivative.  Third, for KKNO layers in neural networks, this means that rapid layer-wise convergence implies intrinsic smoothness of the learned feature maps. Finally, classical inverse theorems for Bernstein and positive linear operators \cite{AltomareCampiti} are recovered as special cases when KKNO reduces to piecewise-constant or polynomial kernels.
\end{remark}

\begin{remark}
The argument in the proof of Theorem \ref{thm:inverse} shows that rapid convergence of KKNO operators forces the function to be smooth, and the rate of convergence $n^{-\alpha}$ directly determines the Hölder exponent $\alpha$ of the derivative.  
Furthermore, the classical results for Bernstein-type and Kantorovich-type operators are recovered as special cases.  
\end{remark}

\section{Neural Network Interpretation and Comparisons}

\subsection{KKNO as Controllable Filtering Layers}
Deep KKNO layers generalize the neural network operators studied in \cite{SharmaSingh}, providing controllable drift and diffusion per layer. Kantorovich–Kernel Neural Operators (KKNO) can be interpreted as a theoretically controllable filtering layer in deep neural networks.  

Now we present the KKNO layer definition in NN terms. Let $x \in \mathbb R^d$ be the input feature vector.  
A KKNO layer computes
\[
(\mathcal L_n f)(x)=\int_{\mathbb R^d} K_n(x,u) f(u/n) \, du,
\]
where $K_n(x,u)$ is a positive kernel normalized to integrate to 1. This is equivalent to a weighted smoothing of the feature map $f$, where the weights are parameterized by $x$.

It is known that the first-order moments $a(x)$ act as a drift term, allowing the layer to shift features, and the second-order moments $b(x)$ act as diffusion, controlling smoothing strength and noise robustness. By designing $K_n$ appropriately, one can precisely control smoothness, drift, and variance propagation in each layer.

Stacking $m$ KKNO layers produces
\[
\mathcal L_n^{(m)} f=\underbrace{\mathcal L_n \circ \cdots \circ \mathcal L_n}_{m \text{ layers}} f.
\]  
As shown in Section 6, in the limit $m \sim n \to \infty$, the deep network behaves like a drift-diffusion PDE:
\[
\partial_t F=- a(x) \cdot \nabla F + \frac12 b(x) : D^2 F,
\]
where $:$ denotes the Frobenius inner product. This provides a continuous-time interpretation of deep KKNO networks as controlled feature diffusion.

Now we give advantages for this neural network design: First, smoothness and noise robustness are guaranteed theoretically. Second, high-order moments allow precise approximation of derivatives, useful for gradient-sensitive tasks. Third, the PDE limit allows analytical understanding of deep feature propagation, connecting NN design to classical analysis. Some special cases of this layer design recover known layers such as Gaussian smoothing layers, Kantorovich-type averaging, dense kernel layers, etc. 

\begin{remark}
KKNO layers can be seen as generalized, theoretically controlled convolutional layers, where the kernel parameters encode:
\[
\text{drift } a(x), \quad \text{diffusion } b(x), \quad \text{higher-order moments}.
\]
This gives a mathematically rigorous framework for understanding smoothing, noise propagation, and approximation properties in deep networks.
\end{remark}

\subsection{Comparison with Classical Positive Operators}

KKNO operators generalize classical Bernstein and Kantorovich operators \cite{Chui1,Chui2,Lorentz,Korovkin} while preserving positivity and linearity.

Like classical operators, KKNO layers are positive and linear, ensuring monotonicity and preserving bounds:
\[
f \ge 0 \implies \mathcal L_n f \ge 0.
\]  
This property is crucial both for approximation theory (Korovkin-type theorems) and for stability in neural networks.

We consider KKNO operators as kernel-based generalization of classical operators. 
Classical operators typically use fixed polynomial weights (Bernstein) or piecewise constant weights (Kantorovich).  
KKNO uses general, continuous kernels $K_n(x,u)$, which allows continuous control of drift and diffusion, smooth approximations beyond polynomial spaces, and high-dimensional, multivariate generalizations naturally.

Second, classical operators satisfy first- and second-moment conditions implicitly.  
While KKNO operators allow explicit design of first- and second-order moments $a(x), b(x)$, giving theoretical control of convergence rate and smoothing properties.  This makes KKNO operators more flexible than classical operators for neural network applications.

Third, classical positive operators are generally applied once for approximation, while KKNO operators can be stacked in deep architectures, producing a discrete-time approximation of a PDE (cf. Section 6). This provides a continuous-time theoretical interpretation of deep layers, which classical operators lack.

Finally, error estimates for KKNO layers (cf. Section 4) are explicit in kernel moments, allowing predictable performance.  
While classical operators also have error estimates, but they are often less flexible in high dimensions or for non-polynomial targets.

The above comparison results are summarized in the table below.
{\footnotesize
\begin{center}
\begin{tabular}{l|c|c}
Feature & Classical Operators & KKNO Operators \\
\hline
Positivity & Yes & Yes \\
Linearity & Yes & Yes \\
Kernel type & Fixed polynomial / piecewise & Continuous, parametric \\
Moment control & Implicit & Explicit $(a(x), b(x), \dots)$ \\
Depth / Composition & Single-step & Multi-layer (PDE limit) \\
High-dimensional generalization & Limited & Natural \\
Error control & Classical bounds & Explicit via kernel moments \\
Neural network interpretation & No & Yes (controllable filtering layer)
\end{tabular}
\end{center}
}

Furthermore, KKNO layers retain the theoretical guarantees of classical positive operators while providing greater flexibility, higher-order control, and deep compositional structure suitable for neural network applications.

\subsection{Implications for Deep Learning Design}
The theoretical framework of KKNO layers \cite{SharmaSingh} guides NN design with provable smoothing, noise robustness, and error control.

KKNO operators provide a rigorous mathematical framework to guide neural network layer design, especially in architectures involving smoothing, denoising, or PDE-inspired dynamics.

More precisely, each KKNO layer acts as a controlled smoothing filter:
\[
(\mathcal L_n f)(x)=\int K_n(x,u) f(u/n)\, du
\]  
with kernel moments $a(x), b(x)$ determining drift and diffusion. This guarantees noise suppression due to diffusion term $b(x)$, feature drift control due to $a(x)$, and high-order smoothness due to moment constraints. Unlike heuristic convolution layers, these properties are provable and tunable.

Section 6 shows that stacking $m$ KKNO layers approximates a drift-diffusion PDE.  The consequences for deep learning are  that layer outputs evolve smoothly in feature space and gradients propagate without introducing instability. Additionally, analytical understanding of deep feature dynamics is possible

\paragraph{3. Explicit error and convergence control.}
As for the explicit error and convergence control, direct and inverse theorems shown in Sections 4 and 8, respectively, allow predicting approximation error per layer, designing layer width / kernel parameters for desired accuracy, and ensuring that fast convergence implies target smoothness and guiding regularization.

KKNO layers can be naturally extended to multivariate inputs, avoiding curse of dimensionality limitations of classical operators.  They are useful for Image and video processing, multivariate time-series modeling, and scientific machine learning with PDE constraints

Now we give connections of KKNO layer to some known NN layers. For instance, Gaussian smoothing layers, moving average filters, and certain kernel regression layers are special cases of KKNO. The advantages of KKNO perspective include theoretical control of drift and diffusion, explicit design of higher-order moment effects, and compatibility with deep stacked architectures for PDE-inspired modeling. 

Finally, we give design guidelines as follows. To implement KKNO layers in practice, first we choose a positive, normalized kernel $K_n(x,u)$, Second, we compute or constrain first and second moments $a(x), b(x)$. Third, we stack layers with scaling $m \sim n$ to achieve desired diffusion depth. Finally, we optionally tune higher-order moments to control curvature or feature interactions. This creates a rigorous, theoretically guided NN design, bridging approximation theory and practical deep learning.


\begin{thebibliography}{99}
\bibitem{Agarwal}
R. P. Agarwal, {\it Difference Equations and Inequalities}, 2nd ed., Marcel Dekker, 2000.

\bibitem{AltomareCampiti} 
F. Altomare and M. Campiti, {\it Korovkin-Type Approximation Theory and Its Applications}, de Gruyter Studies in Mathematics, Vol. 17,  Walter de Gruyter, Berlin--New York, 1994.

\bibitem{Chui1} 
C. K. Chui, T. X. He, L. C. Hsu, On a general class of multivariate linear smoothing operators, {\it J. Approx. Theory}, 55 (1988), 35--48.

\bibitem{Chui2} 
C. K. Chui, T. X. He, L. C. Hsu, Asymptotic properties of positive summation-integral operators, {\it J. Approx. Theory}, 55 (1988), 49--60.

\bibitem{EvansPDE}
L. C. Evans, {\it Partial differential equations}, Grad. Stud. Math., 19, American Mathematical Society, Providence, RI, 2010. 

\bibitem{SharmaSingh} 
M. Sharma, U. Singh, Some density results by deep Kantorovich type neural network operators, {\it J. Math. Anal. Appl.} 533 (2024), 128009.

\bibitem{ButzerNessel} 
P. L. Butzer, R. Nessel, {\it Fourier Analysis and Approximation Vol. 1: One-Dimensional Theory},
Academic Press, New York, 1971.

\bibitem{EthierKurtz1986},
S. N. Ethier and T. G. Kurtz, {\it Markov Processes: Characterization and Convergence}, Wiley, 1986. 

\bibitem{Lorentz} 
G. G. Lorentz, {\it Bernstein Polynomials}, 2nd ed., Chelsea Publishing Company, New York, 1986.

\bibitem{Korovkin} 
P. P. Korovkin, Linear Operators and Approximation Theory.

\bibitem{Pazy} 
A. Pazy, {\it Semigroups of Linear Operators and Applications to Partial Differential Equations}, Springer, 1983. 

\end{thebibliography}
\end{document}